\documentclass{article}

     \usepackage[preprint, nonatbib]{gamma}

\usepackage[utf8]{inputenc} % allow utf-8 input
\usepackage[T1]{fontenc}    % use 8-bit T1 fonts
\usepackage{hyperref}       % hyperlinks
\usepackage{url}            % simple URL typesetting
\usepackage{booktabs}       % professional-quality tables
\usepackage{amsfonts}       % blackboard math symbols
\usepackage{nicefrac}       % compact symbols for 1/2, etc.
\usepackage{microtype}      % microtypography
\usepackage{xcolor}         % colors

\usepackage{graphicx}
\usepackage[english]{babel}
\usepackage{algorithm}
\usepackage[noend]{algpseudocode}
\usepackage{comment}

\title{GAMMA: Generative Augmentation for Attentive Marine Debris Detection}

\author{%
  Vaishnavi Khindkar\\
  \texttt{vkhindkar@gmail.com} \\
   \And
   Janhavi Khindkar \\
   \texttt{jkhindkar14@gmail.com} \\
}

\begin{document}
%\setcitestyle{square}

\maketitle

\begin{abstract}
 We propose an efficient and generative augmentation approach to solve the inadequacy concern of underwater debris data for visual detection. We use cycleGAN as a data augmentation technique to convert openly available, abundant data of terrestrial plastic to underwater-style images. Prior works just focus on augmenting or enhancing existing data, which moreover adds bias to the dataset. Compared to our technique, which devises variation, transforming additional in-air plastic data to the marine background. We also propose a novel architecture for underwater debris detection using an attention mechanism. Our method helps to focus only on relevant instances of the image, thereby enhancing the detector performance, which is highly obliged while detecting the marine debris using Autonomous Underwater Vehicle (AUV). We perform extensive experiments for marine debris detection using our approach. Quantitative and qualitative results demonstrate the potential of our framework that significantly outperforms the state-of-the-art methods.
  % Prior works only focus on image enhancements, while some use variational autoencoders to generate synthetic imagery by feeding random noise to existing data. 
  
\end{abstract}
\section{Introduction}
\label{sec1}
Object detection has shown tremendous improvements in recent years especially after the invention of convolution neural networks which led to numerous innovations in the fields like robotics, medical imaging, autonomous driving. Although, a hurdle in training the object detection models is that they usually rely on a vast amount of training data. This becomes daunting especially while dealing with the marine domain, where collecting data is not just time-consuming but also requires huge planning and a lot of manual intervention. Prior work \cite{fulton2019robotic,hong2020generative,tata2021deepplastic} shows that object detectors trained on such underwater datasets show poor generalization as compared to their terrestrial counterparts due to the data scarcity concern of the marine domain. Moreover, the domain-specific issues which add up to the challenge of underwater object detection and especially marine debris detections are as stated - i. The shape of the object gets changed due to biological degradation and water pressure. ii. Improper light conditions and turbidity add bias and difficulty in detection. iii. Similar looking classes like suspended debris like plastic bags and some of the marine life. iv. Color absorption differs as per depth.  Motivated by these challenges, we aim to solve the problem of marine debris detection, by proposing a generative variational approach to solve the data scarcity issue and by designing a novel and efficient attention-based detection strategy. %These all issues are subproblems of the 2 main challenges in the marine domain namely Data scarcity of proper data and the lack of SOTA object detectors to efficiently work in the underwater domain.

\begin{figure}
  \centering
  %\fbox{\rule[-.5cm]{0cm}{4cm} \rule[-.5cm]{4cm}{0cm}}
  \includegraphics[width=1.0\linewidth]{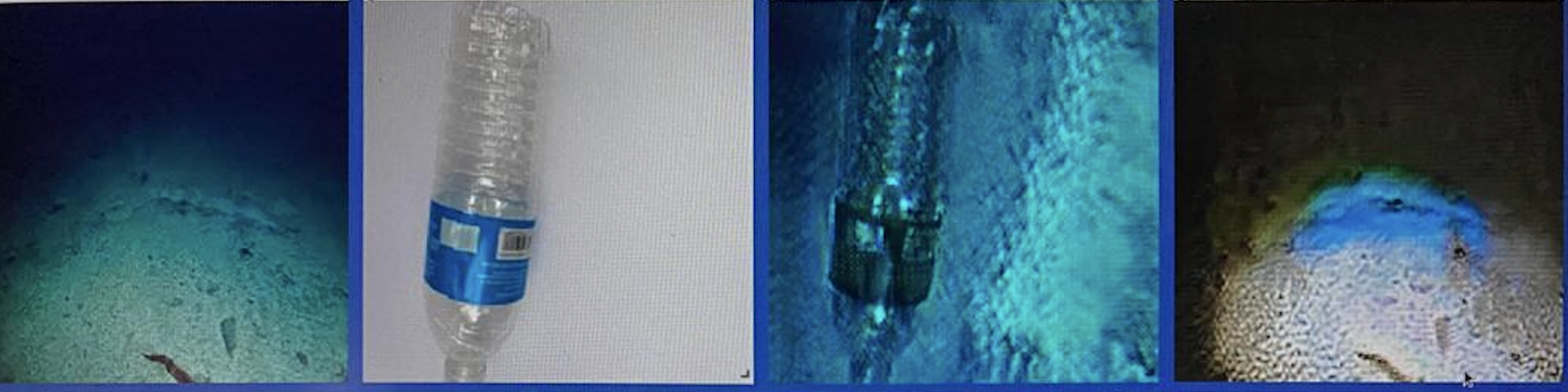}
  \includegraphics[width=1.0\linewidth]{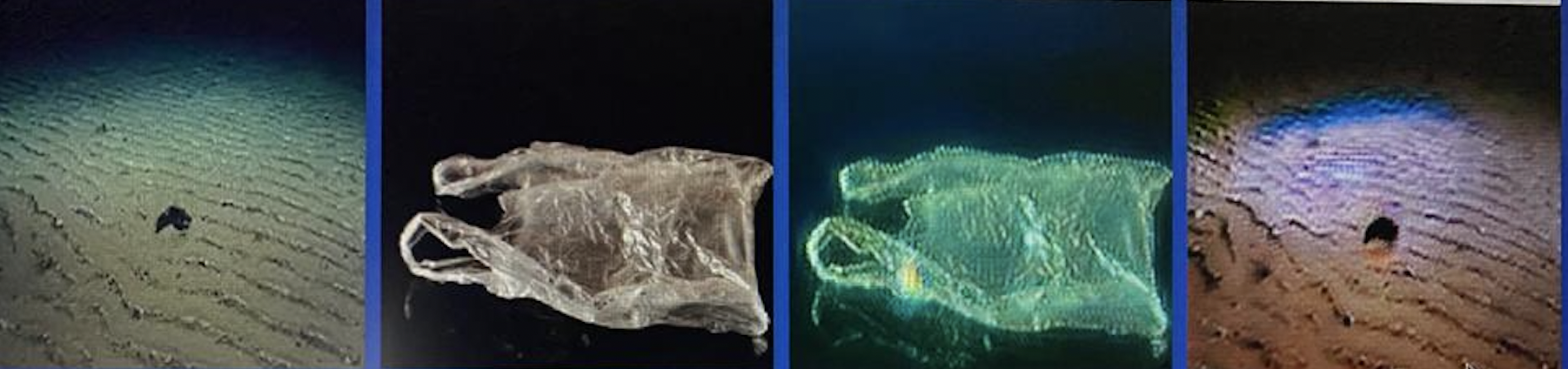}
  \caption{Efficient data augmentation using cycleGAN to convert (in-air) terrestrial plastic images to underwater marine domain, generating variational underwater plastic data.}
  \label{fig1}
\end{figure}

The data scarcity problem can’t be solved using simple augmentations like rotation or translation as it doesn't add much variance and is merely the same image from a different angle. To solve this problem we can make use of easily available rich terrestrial data. The terrestrial data can be mapped to the underwater domain using a mapping function $G : X \rightarrow Y$. G maps the terrestrial data to domain Y such that the generated image is of the underwater domain. This can be achieved with the help of Generative Adversarial Networks (GANs) \cite{goodfellowgenerative}. %Image synthesis has shown impressive results after the development of the Generative Adversarial Networks. 
Over years GANs are used for various image applications like image generation \cite{denton2015deep,radford2015unsupervised}, image editing \cite{zhu2016generative}, and image-to-image translations \cite{isola2017image,emami2020spa,zhu2017unpaired}. Image to Image translation can be used to map one domain image to another domain. Most of the image translation models require paired data of different domains for translation and mapping. But this is not possible in our case, as we cannot have simultaneous paired images for underwater and terrestrial domains of marine debris. The recent architecture of GAN, cycleGAN \cite{zhu2017unpaired} can solve this problem, without the need for paired images. It works with unpaired images of different domains, wherein one image can be terrestrial plastic and the second one can be of underwater style. Moreover cycleGAN use cycle consistency loss which ensures that for any input image $x$, the learned mapping should be able to bring $x$ back to the original image, i.e., $x\rightarrow G(x) \rightarrow F (G(x)) = x$. This ensures that the learned function generates the desired mapping and produces stable results. As shown in figure \ref{fig1}, where we map two different domain images of (a) Marine background and (b) Terrestrial plastic, to generate (c) Efficient underwater plastic data. Image augmentation in this way not only helps us in adding variation to the dataset but is also able to solve the problem of degradation of marine debris and lighting conditions of images.

\graphicspath{ {./images/} }

The object detector fails in conditions where the light visibility is very low and the detector performance completely depends upon the camera conditions. Also, prior work \cite{tata2021deepplastic} shows that the detector fails to classify similar-looking species like jellyfish and a plastic bag due to similar appearances and low visibility. So to identify and detect the instances in such a domain, we require added attention such that the model will be able to enhance and identify the objects irrespective of the background noise and lighting conditions. Considering the issues in the existing methods as discussed above we propose a novel object detection method based on the successful mechanism of self-attention \cite{wang2018non} to be able to classify similar looking classes and to detect objects irrespective of the aforementioned conditions. The motivation to use self-attention is based on the structural importance of the objects to enhance the detector performance. %Self-attention helps to focus on the marine debris instead of the background and helps the model to generalize well. Also, the enhanced structural importance helps to identify the similar-looking classes efficiently.
The main contributions of this work are three-fold:
\begin{itemize}
\item We propose a novel Marine Debris dataset using a Generative approach by using cycleGAN.
\item We believe we are the first to incorporate self-attention in marine debris detection. We propose a novel Generative Augmentation for Attentive MArine debris detection (GAAMA) method which is used for dataset creation and that comprises a novel Self-Attentive Marine Debris Detector (SEA) module. SEA generates self-attention feature maps which help improve marine debris detection by enhancing the structural importance of objects.
\item We conduct extensive experiments on the benchmark J-EDI dataset annotated by the University of Minnesota robotics lab \cite{fulton2019robotic}. The framework significantly boosts the performance of existing Marine debris Faster R-CNN \cite{ren2015faster} ,yolo \cite{redmon2016you}, ssd \cite{liu2016ssd} detectors, achieving 9.3\%, and 11\% improvements in the mAP scores using our dataset w/o SEA and with sea module as compared to the state-of-the-art.
\end{itemize}

\section{Background}
\label{sec2}
Our approach lies at the intersection of multiple fields. In the following, we review related work.

\textbf{Underwater Debris Detection:} Increasing plastic pollution has been one of the biggest problems in this century. This has resulted in a surge of research for identifying and removing plastic from the world’s waterways using computer vision and AUV solutions. Japan Agency for Marine-Earth Science and Technology (JAMSTEC) has made a dataset of deep-sea debris available online as part of the larger J-EDI (JAMSTEC E-library of Deep-sea Images) dataset containing images and videos dating 1982. A team of researchers at the University of Minnesota robotics lab \cite{fulton2019robotic} experimented and compared the performance of the various state-of-the-art models for object detection like Faster R-CNN, SSD, YOLO  using metrics of mAP, IOU, and frame rate with Faster R-CNN performing the best. They also released the annotated dataset of images collected by the Japan Agency for Marine-Earth Science and Technology (JAMSTEC). Deeplastic \cite{tata2021deepplastic} conducted a similar kind of research using the above-mentioned annotated dataset along with dataset from 3 sites across California thus generating a total of 3200 training images on which they trained Yolov4 \cite{bochkovskiy2020yolov4} and Yolov5 \cite{glenn_jocher_2021_4679653} models.

\textbf{Image Data Synthesis}: Over recent years Generative Adversarial Networks \cite{goodfellowgenerative} based image synthesis has gained a lot of popularity. Generative models based augmentation has been used for various non-water domain work to improve object detection and classification tasks. Prior works show the use of GANs for an underwater domain such as WaterGAN \cite{li2017watergan}, UGAN \cite{fabbri2018enhancing}, and  Multi-style GAN (UMGAN) \cite{cao2018recent} for image enhancement purposes to improve detection accuracy. WaterGAN uses pairs of non-underwater images and depth maps to generate underwater images. UGAN uses CycleGAN \cite{zhu2017unpaired} to generate distorted images with the Wasserstein GAN \cite{arjovsky2017wasserstein} used to ensure the stability of output. UMGAN is a combination of CycleGAN and Conditional GAN \cite{mirza2014conditional} to generate multi-style underwater images. With all these methods, it is challenging to realistically simulate color and shape distortions. To solve the generalization problems of the above-mentioned techniques researcher at the University of Minnesota robotics lab \cite{hong2020generative} uses 2 staged VAE \cite{dai2019diagnosing} to generate underwater images and a binary classifier to check whether the image is good or bad. Although this can avoid distortions or using complex architectures as mentioned in the above approaches it results in the generation of blur images which affects the performance of the model. To handle problems of blur and distorted images, we propose a simple CycleGAN based approach to generate underwater domain debris data, translating terrestrial debris to the underwater background. Our approach generates non-distorted and sharpened images which help in improving the detection accuracy.

\textbf{Self Attention}
Attention \cite{vaswani2017attention} was first introduced for the encoder-decoder in a neural sequence transduction model to allow for content-based summarization of information from a variable-length source sentence. Since its introduction, the ability of attention to learning to focus on important regions within a context has made it an important component in natural language models \cite{wu2016google,chorowski2015attention,chan2016listen}. Since then using Self-attention in models has been the primary focus in these models. The ability of self-attention to capture global dependencies and handle parallelization, which leverages the strengths of modern hardware, has led to state-of-the-art models for various tasks. Attention has been popular in computer vision networks as well. It has been used in various tasks like SAGAN \cite{zhang2019self}, semantic segmentation and object detection etc \cite{ramachandranstand,caron2021emerging}. We try to leverage this advantage of self-attention in the marine domain to focus on important structures and objects and diminishing the role of background and the lighting conditions.

\graphicspath{ {./images/} }

\begin{figure}
  \centering
  %\fbox{\rule[-.5cm]{0cm}{4cm} \rule[-.5cm]{4cm}{0cm}}
  \includegraphics[width=1.0\linewidth]{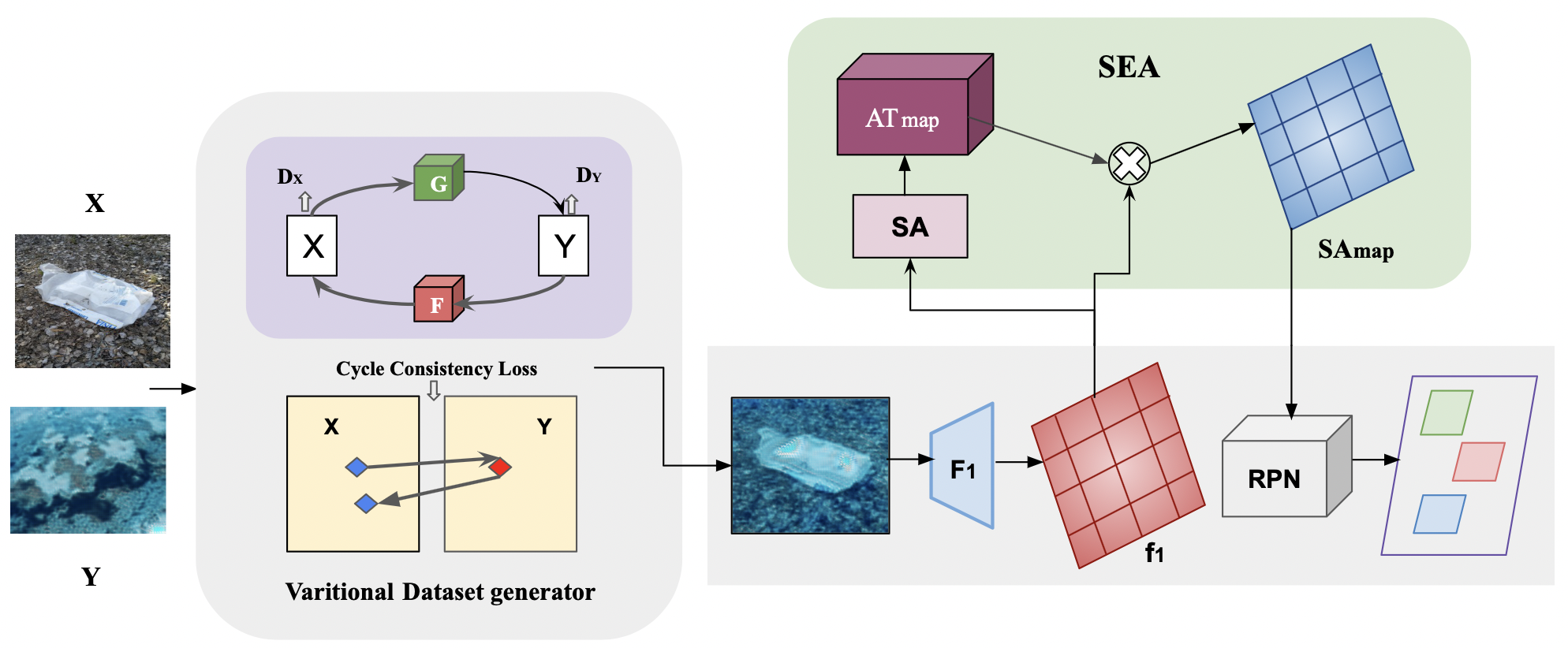}
  \caption{Proposed architecture of GAMMA. It comprises of a variational dataset generator used for augmentation of in-air terrestrial debris to underwater style and novel SEA module that uses self-attention mechanism to efficiently improve the marine debris detection performance.}
  \label{fig2}
\end{figure}

\section{Generative Augmentation for Attentive Marine Debris Detection}
\label{sec3}

In this paper we pose the question: can we effectively solve the problem of marine debris detection by solving the data scarcity issue and improving the detector to effectively detect marine debris irrespective of the background conditions? In doing so, we propose a novel approach for solving the problem: Generative Augmentation for Attentive MArine debris detection (GAAMA).  As shown in figure \ref{fig2}, GAAMA is a two-fold method. The left part of the figure depicts the process of generative dataset creation which is the mapping function used to map the abundant terrestrial debris images to the underwater domain using CycleGAN. Irrespective of the prior works which produce the blurred or distorted images our method produces stable sharpened images without the need for any extra combination with GAN or any need of paired images. This helps to solve the problem of data scarcity and issues with blurred and turbid images.

Prior work in the field uses state-of-the-art object detection models like Faster R-CNN, SSD, and YOLO models without any modification. These models usually fail due to incompatibility to handle varying light conditions and the blurriness induced due to the turbidity of the underwater domain as they process the information in a local neighborhood and are mostly suited for the terrestrial domain. To tackle this problem we propose the novel architecture of SElf Attentive marine debris detector (SEA) as shown in the figure. We make use of self-attention which focuses on the structural importance of the objects thus helping in the detection and classification of the debris irrespective of any conditions induced by the domain.

\subsection{Generative Dataset Creation}

Our key idea is to generate a mapping function $\phi$ that maps $X$ to $Y$ such that 
$\phi : X \rightarrow Y$
where $x_i \in X$ is the set of terrestrial debris given training samples $X_i^n$ and $y_i \in Y$ is the set underwater styles given the samples ${Y_j^m}$, where $n$ and $m$ are the numbers of images for terrestrial and underwater styles respectively. The data distribution of terrestrial debris as  $x$ \textasciitilde  $p_{data}(x)$ and underwater styles as $y$ \textasciitilde $ p_{data}(y)$. The mapping should be such that it does not require paired images as it is really difficult in the underwater domain to have them. So we use cycleGAN to achieve the mapping due to its success with the unpaired images. As a workaround for paired images, cycleGAN uses two mappings $G : X \rightarrow Y$ and $F : Y \rightarrow X$ where $G$ maps terrestrial debris to $X$ to water styles and $F$ maps the output of $G$ back to $X$. To make this happen it uses two discriminators and two generators for each mapping. The adversarial loss is applied on both the mappings. For mapping $G : X \rightarrow Y$ and discriminator $D_Y$ the adversarial loss is defined as,
\begin{equation}
    L_{M_{GAN}}(G, D_Y , X, Y) =  \mathbb{E}_{y \in p_{data}(y)}[log D_Y (y)] + \mathbb{E}_{x \in p_{data}(x)}[log(1 - D_Y (G(x))]
\end{equation}

To ensure the generated underwater images from mapping $G$ does not completely erode the debris in $X$ and the output of $F$ does not contradict input $X$ i.e $y \rightarrow F (y) \rightarrow G(F (y)) = y$ we use cycle consistency loss which is defined as,
\begin{equation}
    L_{cyc}(G, F ) = \mathbb{E}_{x \in p_{data}(x)} [ F (G(x)) - x]+ \mathbb{E}_{y\in p_{data}(y)}[G(F(y)) - y]
\end{equation}

Thus the complete objective by combining the adversarial and cycle consistency loss is given by,
\begin{equation}
    L(G, F, D_X , D_Y ) =L_{M_{GAN}}(G, D_Y , X, Y ) + L_{M_{GAN}}(F,D_X,Y,X)+ \lambda L_{cyc} (G, F )
    \label{eq3}
\end{equation}
                 
\subsubsection{Experiment Setting}

The training set $C={(X_i^m,Y_j^n)}$ where $X_i,Y \in R^{(H\times W \times 3)}$. To generate these we collected both the terrestrial debris and marine style data by web scraping of images and videos. We collected a total of $N=1500$ images of terrestrial debris and $M=500$ of underwater styles. Since gans are subjected to non-stability we ensured the stability of the generated data by conducting various experiments to tune the hyperparameters. 

To evaluate the quality and stability of generated images we use Frechet Inception Distance (FID) score. The lower the score the higher is the quality of generated images. To perform the evaluation it extracts features from both the real images and output of the gans. The data distribution of the  extracted features modeled separately using a multivariate gaussian distribution. Then the FID score is calculated by ${FID(x,g}) = || \mu_x - \mu_g ||_2^2 + T_r(\sigma_x + \sigma_g - 2(\sigma_x \sigma_g)^{1/2} )$
where $\mu$ is mean, and $\sigma$ is covariance.

\subsection{SEA : Self Attentive Marine Debris Detector}

In the case of marine domains, it is very difficult even for humans to detect and then classify the objects into the correct category due to the various problems of light conditions, color similarity, and objects similarity with each other. The water adds more to the problem than the image or video. To solve this problem we can make the object detector focus on the structure of objects instead of the background, which is similar to how we normally detect the objects. Inspired by the structural importance of objects in the scene to improve object detection we proposed a novel architecture of SElf Attentive marine debris detector (SEA). Thus to pay more attention to important structures and to highlight these points in the input image we use the technique of self-attention. Attention is a simple mechanism that makes use of key, query, and value. The key and query decide how much value is important for a particular operation. This value is the attention given to each region in the input. 

Initially, the features are extracted from a batch of training data, consisting of images from training data. These features are then fed to the SElf Attentive marine debris detector module. SEA generates the self-attention maps representing important structures/regions in the feature vector. The map is then concatenated with the feature vector to highlight the important regions which should be focussed by the detector. This output is then passed to the RPN layer to train the Faster R-CNN.

\subsubsection{Experiment Setting}
Input to the detector is labeled source data $D_s = (X_i , Y_i )$ where $X$ represents the marine debris images and $Y \in R^{(m\times 5)}$ represents the list of bounding boxes and corresponding class labels for that particular image. Our proposed method combines self-attention with the Faster R-CNN framework.
Initially the image $x_i \in X$ is passed to the feature extractor $F_1$ as 
$f_1 = F_1(x_i)$.
$F_1$ extracts image-level features $f_1$, which are then passed to the SEA module to get the attention map. The self-attentive feature maps are generated using the self-attention module where we use key, query, and value vectors as discussed above. The $Q$ (query) and $K$ (key) undergoes a matrix multiplication, \begin{equation}
    QK_1 = Q_1 \times K_1
\end{equation}
The output passes through a softmax which converts the resultant vector into a probability distribution, and then it finally gets multiplied by V (value), where $ K,Q,V \in R^{(H\times W\times C)}$.
\begin{equation}
    AT_{map} = softmax(QK)\times V
\end{equation}

To get the final self-attention map the $AT_{map}$ is concatenated with the extracted feature $f_1$ to highlight the structurally important regions in the feature vector.
\begin{equation}
    SA_{map} = AT_{map} \times \gamma + f_1
    \label{eq6}
\end{equation}
This finally is fed to RPN to learn the attentive object detector and generate the final output.

\subsection{Algorithm}
The novel algorithm GAMMA is explained in Algorithm 1. It explains the two main functions of the proposed algorithm namely SEA and GAMMA. The function SEA depicts the steps to calculate self-attention maps of the feature vectors. Function GAMMA is the main backbone network. It first creates the new dataset using Generative Variational mapping of $\phi$. It uses cycleGAN as the mapping. The generated data is then passed to the training phase where the features are extracted and then are fed to the SEA function which generates the self-attention feature maps. And this final self-attention output is fed to the RPN to learn the attentive marine debris detector.
%\begin{algorithm}[H]
%\SetAlgoLined
%\KwResult{Write here the result }
% initialization\;
% \While{While condition}{ 
%  instructions\;
%  \eIf{condition}{
%   instructions1\;
%   instructions2\;
%   }{
%   instructions3\;
%  }
% }
% \caption{How to write algorithms}
%\end{algorithm}
\begin{comment}
\begin{algorithm}[H]
\caption{GAMMA : Generative Augmentation for Attentive Marine Debris Detection}
\SetKwInput{KwInput}{Input}                % Set the Input
\SetKwInput{KwOutput}{Output}              % set the Output
\DontPrintSemicolon
  
  \KwInput{Dataset of terrestrial images and underwater styles}
  \KwOutput{Ldet}

% Set Function Names
  \SetKwFunction{FMain}{GAMMA}
  \SetKwFunction{FSEA}{SEA}
 
% Write Function with word ``Function''
  \SetKwProg{Fn}{Function}{:}{}
  \Fn{\FSEA{$f1$, $Q$, $K$, $V$}}{
        QK = softmax(Q * K)\;
        ATmap = QK × V		\;
        SAmap = AT ×f1	\;
        \KwRet SAmap\;
  }
  \;

  \SetKwProg{Fn}{Function}{:}{\KwRet}
  \Fn{\FMain}{
        a = 5\;
        b = 10\;
        Sum(5, 10)\;
        Sub(5, 10)\;
        print Sum, Sub\;
        \KwRet 0\;
  }
 
\end{algorithm}

\begin{algorithm}
\caption{GAMMA : Generative Augmentation for Attentive Marine Debris Detection}
\begin{algorithmic}
\STATE $T\gets loadTerrrestrial()$
\STATE $W\gets loadMarineStyle()$  \COMMENT{a}
\STATE $Data\gets \phi(T,W)$
\FOR{$x_i$, $ y_i$ $\in$ $ Data$ }
    \STATE $f1\gets F1(x_i)$
    \STATE $SAmap\gets SEA(f1,Q,K,V)$
    \STATE $Ldet\gets RPN(SAmap,y_i)$
\ENDFOR
\end{algorithmic}
\end{algorithm}
\end{comment}

\begin{algorithm}
\caption{GAMMA : Generative Augmentation for Attentive Marine Debris Detection}
\begin{algorithmic}[1]
\vspace{1mm}
\Function{SEA}{\textit{$f_1$, Q, K, V}}       %\Comment{Function to calculate Self Attention Maps}
    \State $QK \leftarrow softmax(Q \times K)$ \Comment{Converts the resultant vector into probability distribution}
    \State $ATmap \leftarrow QK \times V$ \Comment{ Generates attention maps}
    \State $SAmap \leftarrow ATmap \times \gamma + f_1$\Comment{ Concatenates attention maps with base feature map.}
    \State $\textbf{return } SAmap$
\EndFunction

\vspace{3mm}

\Function{GAMMA}{\textit{$\phi,F_1,Q,K,V$}}       %\Comment{Algorithm of new proposed method}
    \State $T \leftarrow loadTerrestrialData()$ %\Comment{Load terrestrial debris dataset}
    \State $W\gets loadMarineStyleData()$  %\Comment{Load Marine style dataset}
    \State $Data\gets \phi(T, W)$    \Comment{Marine debris dataset generator using $\phi$ (cycleGAN)}
    \For{$x_i$, $ y_i$ $\in$ $ Data$ }\Comment{Training Attentive Object Detector }
        \State $f_1\gets F_1(x_i)$    %\Comment{Get feature Map}
        \State $SAmap\gets SEA(f1,Q,K,V)$ \Comment{Generate Self-attention map}
        \State $Ldet\gets RPN(SAmap,y_i)$ \Comment{Generate ROIs and detection loss}
    \EndFor
    \State \textbf{return } Ldet
\EndFunction

\end{algorithmic}
\end{algorithm}

\section{Experiments}
\label{sec4}

\subsection{Implementation Details}   %training details, graphs, evaluation metrics
We use the PyTorch framework for all the training tasks. Our proposed dataset consists of a total of $7,930$ images of which $6,786$ images are used for training and $1,144$ images are used for evaluation. We use a $60-40\%$ split for existing (J-EDI) and augmented training data using our dataset generator framework, which is only used in training, evaluation is done only on non-augmented test images. 

\textbf{Variational dataset generator using cycleGAN.} Following the practices in \cite{zhu2017unpaired}, we train our network from scratch, with a learning rate of $0.0002$. Similarly, we divide the objective by two while optimizing the discriminator, which slows down the rate at which it learns, relative to the rate of the generator. We keep the same learning rate for the first $100$ epochs and linearly decay the rate to zero over the next $100$ epochs. Weights are initialized from a gaussian distribution $N(0,0.02)$. For all the experiments, we set $\lambda = 10$ in equation \ref{eq3}. We use the Adam solver \cite{DBLP:journals/corr/KingmaB14} with a batch size of $1$. We train the model for over $200$ epochs.

\textbf{Attentive Object Detection.} We employ VGG-16 \cite{DBLP:journals/corr/SimonyanZ14a} as backbone network for detection, where their weights are pre-trained on the Imagenet Dataset \cite{5206848}. In all our experiments the shorter side of all the training and testing images is resized to 600. Each batch in our training data is composed of four images. We fine-tune the detection network with a learning rate of $1 \times 10^{-3}$ for $1600$ iterations and then reduce the learning rate to $1 \times 10^{-4}$ for the other $2000$ iterations. The momentum of $0.9$ and the weight decay of $5 \times 10^{-4}$ is used for VGG-16 \cite{DBLP:journals/corr/SimonyanZ14a} based detectors. In all experiments, we use RoI-Align \cite{He_2017_ICCV} for RoI feature extraction. For all the experiments, we use mean average precision (mAP) metrics to evaluate the detection performance.

\subsection{Memory Usage}
All the models are trained using a Linux Machine, using an NVIDIA GTX $1060$ GPU. Training of the presented models takes on the order of 2 days for variational data generator and around a day for the attentive object detector. A single forward pass takes around 6 GB of GPU memory per batch item for an attentive object detector.

\subsection{Results}  %tables, qualitative results, failure cases
Table \ref{tab:my_label1} shows our experimental results for the task of marine debris detection compared to the state-of-the-art as well as w or w/o using the SEA module. We can see that our model beats the state-of-the-art \cite{fulton2019robotic} detection results, with an improved performance gap of over $11\%$ using the SEA module and over $9.3\%$ w/o using our SEA module. Also, we can see significant improvement in per class detection APs, as over $12.3 \%, 17.1 \%$ and $21.8 \%$ of the plastic, Rov, and bio classes respectively, compared to the state-of-the-art \cite{fulton2019robotic}. The per-class APs of GAMMA w/o using the SEA module are also significantly increased, which shows the potential of our variational dataset generator. We get over $3 \%$ improvement in the overall mAP of marine debris detection using our SEA module, which proves the efficiency of our proposed architecture using the self-attention mechanism. Also, it can be seen that we get significant improvement in the mAP of the plastic class compared to \cite{tata2021deepplastic}, although being trained on 3 different marine debris classes, as compared to \cite{tata2021deepplastic} that only trains on plastic class for marine debris detection. Notably, it's also because of the improved and efficient augmentation in our dataset as compared to \cite{hong2020generative}.

\begin{table} [t]
\centering
%\resizebox{\columnwidth}{!}
%\resizebox{\textwidth}{!}
%\setlength{\tabcolsep}{2pt}
%\footnotesize
\caption{Comparative results for marine debris detection}
\begin{tabular}{c | c | c c c | c}
\hline
Method & Dataset & Plastic & Rov & Bio & mAP \\
\hline
Trash Detector \cite{fulton2019robotic} & JAMSTEC JEDI & 83.3 & 73.2 &71.3 & 81.0\\
Deep Plastic \cite{tata2021deepplastic} & Deep Trash (Custom) & 85.0 & - & - & 85.0 \\
GAMMA w/o SEA & GAMMA & 90.8 & 89.8 & 90.6 & \textbf{90.3}\\
\textbf{GAMMA (Ours)} & GAMMA & 95.6 & 90.3 & 93.1 & \textbf{93.0}\\
\hline
\end{tabular}
\label{tab:my_label1}
\end{table}

\graphicspath{ {./images/} }

\begin{figure}
  \centering
  %\fbox{\rule[-.5cm]{0cm}{4cm} \rule[-.5cm]{4cm}{0cm}}
  \includegraphics[width=1.0\linewidth]{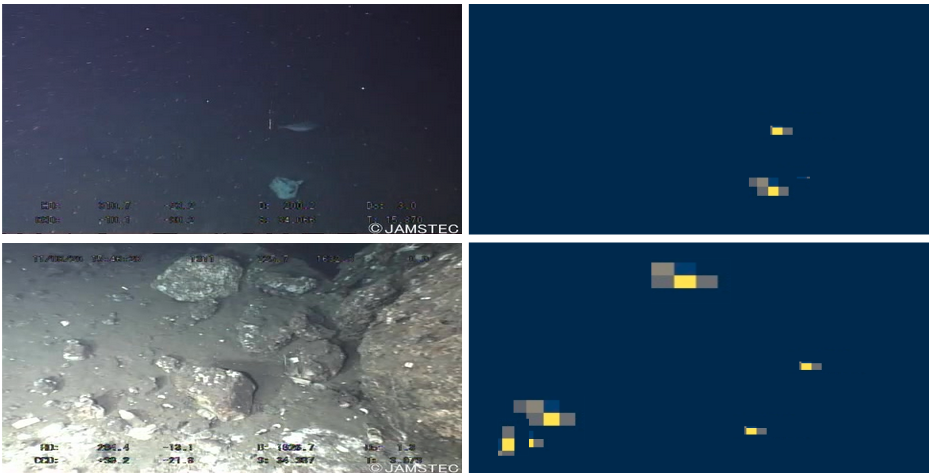}
  \caption{Attention map visualisation using SEA module}
  \label{fig3}
\end{figure}

\graphicspath{ {./images/} }
%\vspace{10mm}
\begin{figure}
  \centering
  %\fbox{\rule[-.5cm]{0cm}{4cm} \rule[-.5cm]{4cm}{0cm}}
  \includegraphics[width=1.0\linewidth]{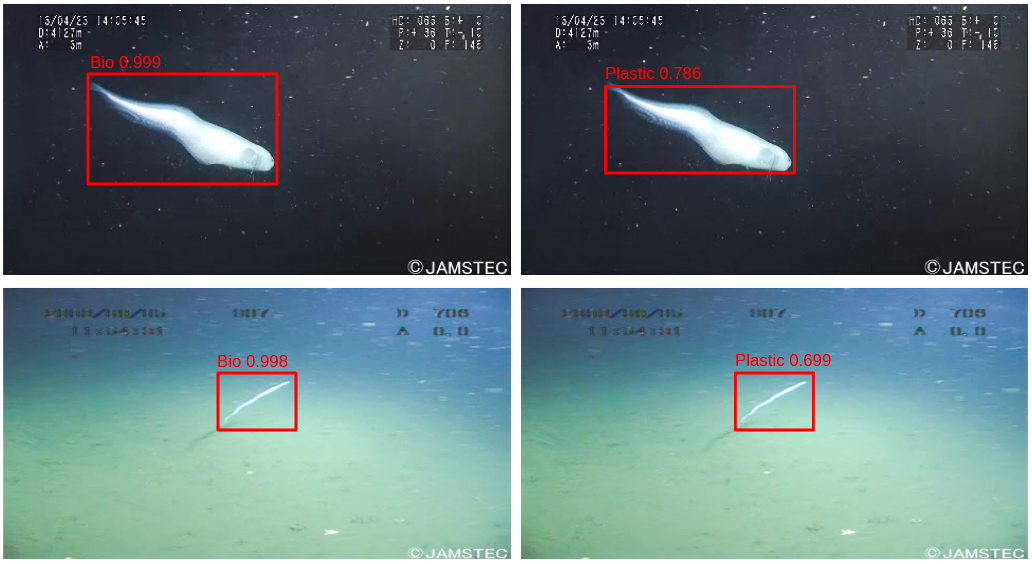}
  \includegraphics[width=1.0\linewidth]{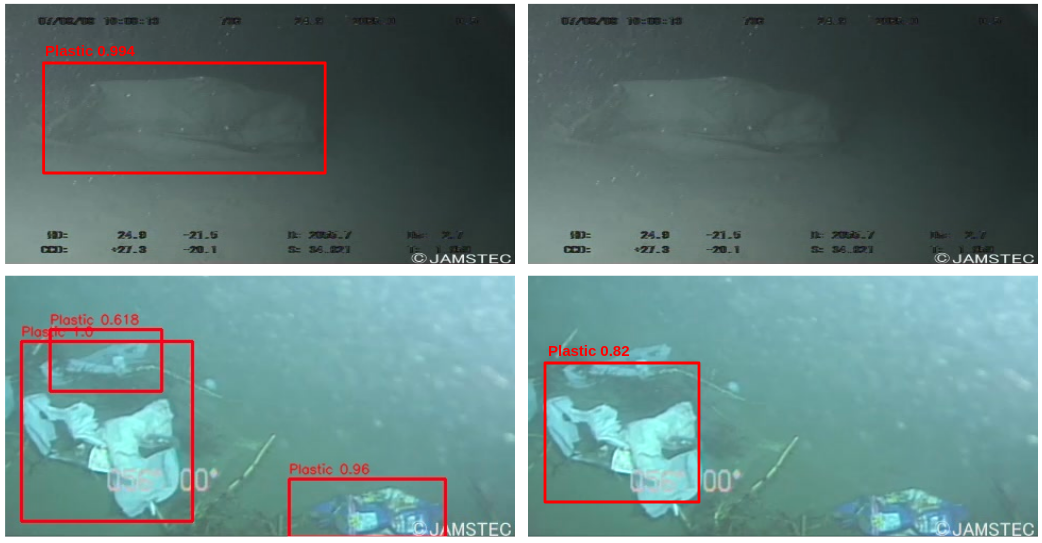}
  \includegraphics[width=1.0\linewidth]{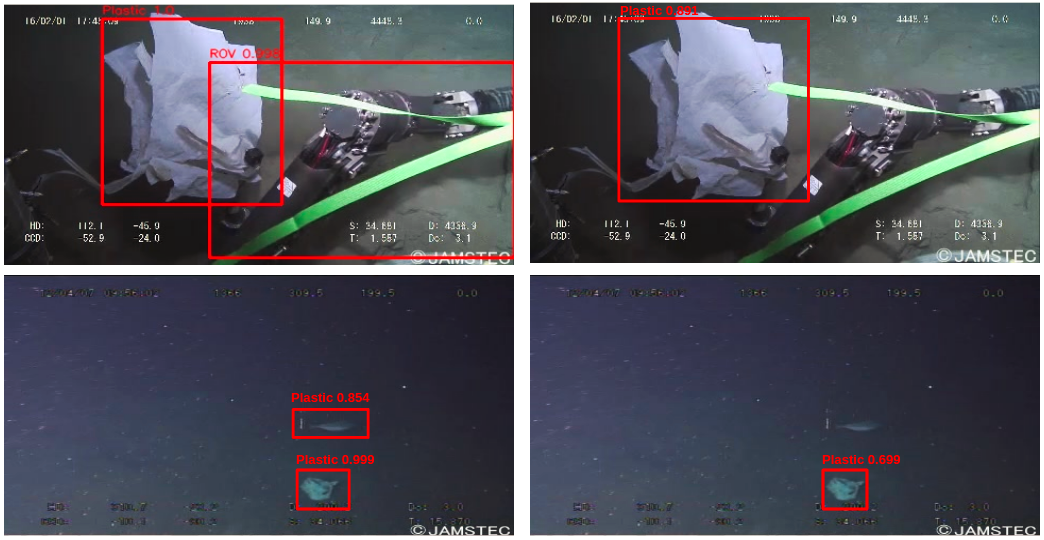}
  \vspace{-5mm}
  \caption{Marine Debris Detection results using GAMMA (left) and the state-of-the-art (right).}
  \label{fig4}
\end{figure}

\textbf{Qualitative Visualisation and Analysis of Marine Debris Detection Results.} We perform qualitative visualizations and analysis on the marine debris detection results. As seen in figure \ref{fig4}, the comparative detection results of our proposed method GAMMA (left) and the state-of-the-art (right) are shown. We can see that our method performs significantly improved detection as compared to the state-of-the-art \cite{fulton2019robotic} and most importantly all the issues that we discussed regarding the marine debris detection regarding turbidity or similar-looking classes etc are all resolved using our method. As shown in the figure, the first two rows show the cases where our model, GAMMA detects and classifies the Bio class with high accuracy as compared to the state-of-the-art method that detects it as a plastic class. This signifies the importance and potential of our method to be able to classify the similar-looking classes correctly without any confusion. The third and fourth rows show the qualitative detection results where our model correctly detects the plastic in the conditions of turbidity as well as the suspended plastic as compared to the state-of-the-art which fails in such cases. Also, the last row shows the detection results where our model detects the hardly observable plastic efficiently where the state-of-the-art method fails as well. All these results simply signify the importance of our proposed method GAMMA and also the SEA module which helps in enhancing the important object instances using attention mechanism, which helps the model successfully detect and classify the marine debris efficiently even in the cases of turbidity, suspension, improper lighting or even similar-looking classes detection. 

\textbf{Visualisation of Attention maps using SEA module.} As shown in figure \ref{fig3}, we visualize the generated self-attention feature maps using the SEA module in our proposed design. These are the output features, $SA_{map}$ as given in the equation \ref{eq6} which are obtained using self-attention mechanism and are further fed to RPN to learn attentive object detector. We can see in figure \ref{fig3}, the attention map visualizations correctly enhance the significant instances of the image, as both the plastic bags in the first row. This further helps in improving detection results as can be seen comparatively in figure \ref{fig4} (last row) as discussed above. \textbf{Failure cases.} Also we show the failure cases, as can be seen in the second row of the figure, where the attention map also tries to attend regions of the image where non-debris objects like rocks come into the picture along with the plastic debris. These are failure cases and we would consider such cases as a part of future work where we work on improving the design using multi-headed attention or additional improvements as required.

\section{Motivation}
\label{sec5}

According to the 5 Gyres organization, nearly 270,000 metric tons and 5.25 trillion items of plastic are on the surface of the world's oceans. The quantification of the buoyant marine plastic debris is important in understanding the concentrations of trash across the world and identifying high concentration garbage hotspots for its smooth removal. Marine plastic debris affects at least 267 species worldwide including 87 percent of all sea turtles and 44 percent of all seabird species and is also a cause of various marine animal deaths. It also has a huge impact on wildlife causing ingestion, entanglement, starvation, suffocation, infection, and drowning. Plastics also routinely explode into the water, degrading the water quality with toxic compounds and ends up harming human and animal health. Thus plastic pollution not only poses an imminent threat to the marine environment but to human health as well. It also has a huge impact on climate change. This makes it one of the biggest problems of this generation and thus requires an efficient and effective solution to tackle it.

Current methods of marine debris cleaning are confined to the surface debris and don't focus much on the suspended and settled plastic at the bed. Currently, the technique to quantify floating plastic requires the use of a manta trawl with manual intervention. This adds difficulty and requires a lot of planning and incurs high cost. To solve this problem we propose a solution using the power of computer vision and deep learning which can be incorporated with autonomous underwater vehicles and help in the detection and collection of marine debris.

\section{Conclusion and Future Work}
\label{sec6}

We presented GAMMA, a novel variational augmentation and attention-based approach for marine debris detection. The relatively strong performance demonstrates the efficiency of our method SEA to detect the debris irrespective of the blurred background and lighting conditions. We also solve the problem of hard-to-detect as well as similar-looking classes and improve the detection performance using an attention mechanism as compared to the state-of-the-art model. We also propose a new marine debris dataset using variational augmentation which beats the state-of-the-art without even the SEA module, highlighting the importance of our variational augmented dataset. Extensive results and visualization prove the potential of the proposed method to perform efficient marine debris detection, which can be also extended for other domains.

In the future, we will try to optimize and quantize the algorithm to deploy it on the edge devices as a part of AUV and test the proposed algorithm in real-time. We will also try to improvise the dataset as well as the SEA (attention) module to handle the failure cases.

\section{Acknowledgments}
The images in the dataset were a combination of the newly created dataset and images sourced from the TrashCan dataset, which was hand-annotated by researchers of the University of Minnesota from the JAMSTEC-JEDI dataset and open-sourced.
The authors would like to thank the researchers and would also like to thank the Japan Agency for Marine-Earth Science and Technology for open sourcing this data.

%\section{References}
{\small
\bibliography{gamma}

\begin{thebibliography}{10}

\bibitem{fulton2019robotic}
Michael Fulton, Jungseok Hong, Md~Jahidul Islam, and Junaed Sattar.
\newblock Robotic detection of marine litter using deep visual detection
  models.
\newblock In {\em 2019 International Conference on Robotics and Automation
  (ICRA)}, pages 5752--5758. IEEE, 2019.

\bibitem{hong2020generative}
Jungseok Hong, Michael Fulton, and Junaed Sattar.
\newblock A generative approach towards improved robotic detection of marine
  litter.
\newblock In {\em 2020 IEEE International Conference on Robotics and Automation
  (ICRA)}, pages 10525--10531. IEEE, 2020.

\bibitem{tata2021deepplastic}
Gautam Tata, Sarah-Jeanne Royer, Olivier Poirion, and Jay Lowe.
\newblock Deepplastic: A novel approach to detecting epipelagic bound plastic
  using deep visual models.
\newblock {\em arXiv preprint arXiv:2105.01882}, 2021.

\bibitem{goodfellowgenerative}
Ian~J Goodfellow, Jean Pouget-Abadie, Mehdi Mirza, Bing Xu, David Warde-Farley,
  Sherjil Ozair, Aaron Courville, and Yoshua Bengio.
\newblock Generative adversarial nets.
\newblock {\em NIPS}, 2014.

\bibitem{denton2015deep}
Emily~L Denton, Soumith Chintala, Arthur Szlam, and Rob Fergus.
\newblock Deep generative image models using a laplacian pyramid of adversarial
  networks.
\newblock In {\em NIPS}, 2015.

\bibitem{radford2015unsupervised}
Alec Radford, Luke Metz, and Soumith Chintala.
\newblock Unsupervised representation learning with deep convolutional
  generative adversarial networks.
\newblock {\em In ICLR}, 2016.

\bibitem{zhu2016generative}
Jun-Yan Zhu, Philipp Kr{\"a}henb{\"u}hl, Eli Shechtman, and Alexei~A Efros.
\newblock Generative visual manipulation on the natural image manifold.
\newblock In {\em European conference on computer vision}, pages 597--613.
  Springer, 2016.

\bibitem{isola2017image}
Phillip Isola, Jun-Yan Zhu, Tinghui Zhou, and Alexei~A Efros.
\newblock Image-to-image translation with conditional adversarial networks.
\newblock In {\em Proceedings of the IEEE conference on computer vision and
  pattern recognition}, pages 1125--1134, 2017.

\bibitem{emami2020spa}
Hajar Emami, Majid~Moradi Aliabadi, Ming Dong, and Ratna~Babu Chinnam.
\newblock Spa-gan: Spatial attention gan for image-to-image translation.
\newblock {\em IEEE Transactions on Multimedia}, 23:391--401, 2020.

\bibitem{zhu2017unpaired}
Jun-Yan Zhu, Taesung Park, Phillip Isola, and Alexei~A Efros.
\newblock Unpaired image-to-image translation using cycle-consistent
  adversarial networks.
\newblock In {\em Proceedings of the IEEE international conference on computer
  vision}, pages 2223--2232, 2017.

\bibitem{wang2018non}
Xiaolong Wang, Ross Girshick, Abhinav Gupta, and Kaiming He.
\newblock Non-local neural networks.
\newblock In {\em Proceedings of the IEEE conference on computer vision and
  pattern recognition}, pages 7794--7803, 2018.

\bibitem{ren2015faster}
Shaoqing Ren, Kaiming He, Ross Girshick, and Jian Sun.
\newblock Faster r-cnn: Towards real-time object detection with region proposal
  networks.
\newblock {\em NIPS}, 2015.

\bibitem{redmon2016you}
Joseph Redmon, Santosh Divvala, Ross Girshick, and Ali Farhadi.
\newblock You only look once: Unified, real-time object detection.
\newblock In {\em Proceedings of the IEEE conference on computer vision and
  pattern recognition}, pages 779--788, 2016.

\bibitem{liu2016ssd}
Wei Liu, Dragomir Anguelov, Dumitru Erhan, Christian Szegedy, Scott Reed,
  Cheng-Yang Fu, and Alexander~C Berg.
\newblock Ssd: Single shot multibox detector.
\newblock In {\em European conference on computer vision}, pages 21--37.
  Springer, 2016.

\bibitem{bochkovskiy2020yolov4}
Alexey Bochkovskiy, Chien-Yao Wang, and Hong-Yuan~Mark Liao.
\newblock Yolov4: Optimal speed and accuracy of object detection.
\newblock {\em arXiv preprint arXiv:2004.10934}, 2020.

\bibitem{glenn_jocher_2021_4679653}
Glenn Jocher, Alex Stoken, Jirka Borovec, NanoCode012, Ayush Chaurasia, TaoXie,
  Liu Changyu, Abhiram V, Laughing, tkianai, yxNONG, Adam Hogan,
  lorenzomammana, AlexWang1900, Jan Hajek, Laurentiu Diaconu, Marc, Yonghye
  Kwon, oleg, wanghaoyang0106, Yann Defretin, Aditya Lohia, ml5ah, Ben Milanko,
  Benjamin Fineran, Daniel Khromov, Ding Yiwei, Doug, Durgesh, and Francisco
  Ingham.
\newblock {ultralytics/yolov5: v5.0 - YOLOv5-P6 1280 models, AWS, Supervise.ly
  and YouTube integrations}, April 2021.

\bibitem{li2017watergan}
Jie Li, Katherine~A Skinner, Ryan~M Eustice, and Matthew Johnson-Roberson.
\newblock Watergan: Unsupervised generative network to enable real-time color
  correction of monocular underwater images.
\newblock {\em IEEE Robotics and Automation letters}, 3(1):387--394, 2017.

\bibitem{fabbri2018enhancing}
Cameron Fabbri, Md~Jahidul Islam, and Junaed Sattar.
\newblock Enhancing underwater imagery using generative adversarial networks.
\newblock In {\em 2018 IEEE International Conference on Robotics and Automation
  (ICRA)}, pages 7159--7165. IEEE, 2018.

\bibitem{cao2018recent}
Yang-Jie Cao, Li-Li Jia, Yong-Xia Chen, Nan Lin, Cong Yang, Bo~Zhang, Zhi Liu,
  Xue-Xiang Li, and Hong-Hua Dai.
\newblock Recent advances of generative adversarial networks in computer
  vision.
\newblock {\em IEEE Access}, 7:14985--15006, 2018.

\bibitem{arjovsky2017wasserstein}
Martin Arjovsky, Soumith Chintala, and Léon Bottou.
\newblock Wasserstein gan, Proceedings of the 34th International Conference on
  Machine Learning 2017.

\bibitem{mirza2014conditional}
Mehdi Mirza and Simon Osindero.
\newblock Conditional generative adversarial nets, arXiv preprint
  arXiv:1411.1784 2014.

\bibitem{dai2019diagnosing}
Bin Dai and David Wipf.
\newblock Diagnosing and enhancing vae models.
\newblock {\em arXiv preprint arXiv:1903.05789}, 2019.

\bibitem{vaswani2017attention}
Ashish Vaswani, Noam Shazeer, Niki Parmar, Jakob Uszkoreit, Llion Jones,
  Aidan~N Gomez, Lukasz Kaiser, and Illia Polosukhin.
\newblock Attention is all you need.
\newblock {\em NIPS}, 2017.

\bibitem{wu2016google}
Yonghui Wu, Mike Schuster, Zhifeng Chen, Quoc~V Le, Mohammad Norouzi, Wolfgang
  Macherey, Maxim Krikun, Yuan Cao, Qin Gao, Klaus Macherey, et~al.
\newblock Google's neural machine translation system: Bridging the gap between
  human and machine translation.
\newblock {\em arXiv preprint arXiv:1609.08144}, 2016.

\bibitem{chorowski2015attention}
Jan Chorowski, Dzmitry Bahdanau, Dmitriy Serdyuk, Kyunghyun Cho, and Yoshua
  Bengio.
\newblock Attention-based models for speech recognition.
\newblock {\em Advances in Neural Information Processing Systems}, 2015.

\bibitem{chan2016listen}
William Chan, Navdeep Jaitly, Quoc Le, and Oriol Vinyals.
\newblock Listen, attend and spell: A neural network for large vocabulary
  conversational speech recognition.
\newblock In {\em 2016 IEEE International Conference on Acoustics, Speech and
  Signal Processing (ICASSP)}, pages 4960--4964. IEEE, 2016.

\bibitem{zhang2019self}
Han Zhang, Ian Goodfellow, Dimitris Metaxas, and Augustus Odena.
\newblock Self-attention generative adversarial networks.
\newblock In {\em International conference on machine learning}, pages
  7354--7363. PMLR, 2019.

\bibitem{ramachandranstand}
Prajit Ramachandran, Niki Parmar, Ashish Vaswani, Irwan Bello, Anselm Levskaya,
  and Jonathon Shlens.
\newblock Stand-alone self-attention in vision models.
\newblock {\em NIPS}, 2019.

\bibitem{caron2021emerging}
Mathilde Caron, Hugo Touvron, Ishan Misra, Herv{\'e} J{\'e}gou, Julien Mairal,
  Piotr Bojanowski, and Armand Joulin.
\newblock Emerging properties in self-supervised vision transformers.
\newblock {\em arXiv preprint arXiv:2104.14294}, 2021.

\bibitem{DBLP:journals/corr/KingmaB14}
Diederik~P. Kingma and Jimmy Ba.
\newblock Adam: {A} method for stochastic optimization.
\newblock In Yoshua Bengio and Yann LeCun, editors, {\em 3rd International
  Conference on Learning Representations, {ICLR} 2015, San Diego, CA, USA, May
  7-9, 2015, Conference Track Proceedings}, 2015.

\bibitem{DBLP:journals/corr/SimonyanZ14a}
Karen Simonyan and Andrew Zisserman.
\newblock Very {D}eep {C}onvolutional {N}etworks for {L}arge-{S}cale {I}mage
  {R}ecognition.
\newblock In {\em 3rd International Conference on Learning Representations,
  {ICLR} 2015, May 7-9, 2015, Conference Track Proceedings}, 2015.

\bibitem{5206848}
J.~{Deng}, W.~{Dong}, R.~{Socher}, L.~{Li}, {Kai Li}, and {Li Fei-Fei}.
\newblock Imagenet: {A} {L}arge-{S}cale {H}ierarchical {I}mage {D}atabase.
\newblock In {\em IEEE Conference on Computer Vision and Pattern Recognition},
  pages 248--255, 2009.

\bibitem{He_2017_ICCV}
Kaiming He, Georgia Gkioxari, Piotr Dollar, and Ross Girshick.
\newblock Mask {R-CNN}.
\newblock In {\em Proceedings of the IEEE International Conference on Computer
  Vision (ICCV)}, Oct 2017.

\end{thebibliography}
%\setcitestyle{square}
\bibliographystyle{unsrt}

}

\end{document}